\documentclass[letterpaper, 10 pt, conference]{ieeeconf}  

\IEEEoverridecommandlockouts                              

\usepackage{amsmath,amssymb,amsfonts}
\usepackage{algorithmic,algorithm}
\usepackage{xcolor}
\usepackage{url}
\usepackage{siunitx}
\usepackage{graphicx}
\usepackage{caption}
\usepackage{subcaption}
\usepackage{float}
\usepackage{hyperref}
\usepackage{svg}
\usepackage{dblfloatfix}
\usepackage{multirow}
\usepackage{cite}
\graphicspath{ {./Figures/} }

\title{\LARGE \bf

Belief Scene Graphs: Expanding Partial Scenes with Objects through Computation of Expectation
}

\author{Mario A.V. Saucedo, Akash Patel, Akshit Saradagi, Christoforos Kanellakis and George Nikolakopoulos
\thanks{The authors are with Robotics \& AI Team, Department of Computer, Electrical and Space Engineering, Lule\r{a} University of Technology, Lule\r{a} SE-97187, Sweden. 
        {Corresponding author: \tt\small marval@ltu.se}}
\thanks{This work has been partially funded by the European Union's Horizon Europe Research and Innovation Programme, under the Grant Agreement No. 101119774 SPEAR.}
}

\begin{document}

\maketitle
\thispagestyle{empty}
\pagestyle{empty}

\begin{abstract}
In this article, we propose the novel concept of Belief Scene Graphs, which are utility-driven extensions of partial 3D scene graphs, that enable efficient high-level task planning with partial information. We propose a graph-based learning methodology for the computation of belief (also referred to as expectation) on any given 3D scene graph, which is then used to strategically add new nodes (referred to as blind nodes) that are relevant to a robotic mission. We propose the method of Computation of Expectation based on Correlation Information (CECI), to reasonably approximate real Belief/Expectation, by learning histograms from available training data. 
A novel Graph Convolutional Neural Network (GCN) model is developed, to learn CECI 
from a repository of 3D scene graphs. As no database of 3D scene graphs exists for the training of the novel CECI model, we present a novel methodology for generating a 3D scene graph dataset based on semantically annotated real-life 3D spaces. 
The generated dataset is then utilized to train the proposed CECI model and for extensive validation of the proposed method. We establish the novel concept of \textit{Belief Scene Graphs} (BSG), as a core component to integrate expectations into abstract representations. This new concept is an evolution of the classical 3D scene graph concept and aims to enable high-level reasoning for task planning and optimization of a variety of robotics missions. 
The efficacy of the overall framework has been evaluated in an object search scenario, and has also been tested in a real-life experiment to emulate human common sense of unseen-objects. 

For a video of the article, showcasing the experimental demonstration, please refer to the following link: \url{https://youtu.be/hsGlSCa12iY}
\end{abstract}


\section{INTRODUCTION}
The growing need to develop high-level autonomy for robotic systems 
has set a broad range of challenges for different focus areas in the robotics research community. One of the most crucial challenge is the development of robotic capabilities that will enable the robots to plan their actions dynamically in the presence of uncertainty. Recently proposed planners incorporate the concept of belief for the task of navigation~\cite{ginting2023}, variability estimation~\cite{Looper2023} and~\cite{Giuliari2022} localization of objects. 
%
However, this exciting research area is still in its initial stage and a majority of such robotic missions 
lack the capability to reason and extract higher-level information from the scene, based on partial and uncertain information, as is the case with belief, also referred to as expectation in the rest of this article.

In order to illustrate the utility of expectation in the context of robotic mission planning, let us consider the following example: if we were to leave a human and a robot in a completely unexplored area and ask them to find a specific object, the expected behavior from the robot is to extensively look for such an object, while exploring the totality of the building. On the other hand, a human would take a fast look into each of the rooms to search for the one that is most likely to contain the object. Moreover, 
a human may also look for the respective rooms in specific areas of the building, 
despite not having fully explored the building. The reasoning behind this decision is based on expectation and could be expressed as: \textit{I need to find a pillow, that is likely to be in the bedroom, the bedroom is probably on the second floor, so I must look for the stairs, which may be close to the entrance door}. 

\begin{figure*}[!ht]
        \centering
        \includegraphics[width=\textwidth]{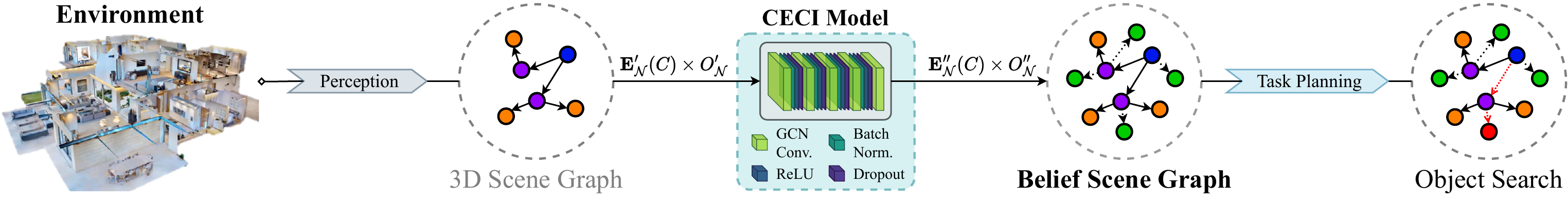}
        \caption{Depiction of a high-level task planning process based on \textit{Belief Scene Graphs}, where the environment is first abstracted into data through perception, in order to build a 3D scene graph representing \textcolor[HTML]{0018EC}{building}, \textcolor[HTML]{9702FA}{rooms} and \textcolor[HTML]{FF8000}{objects}, and then transform into the set of histograms $E^{\prime}_\mathcal{N}(C) \times O^{\prime}_\mathcal{N}$ that are input to the proposed CECI model to output the predicted set of histograms $E^{\prime\prime}_\mathcal{N}(C) \times O^{\prime\prime}_\mathcal{N}$ used to build a \textit{Belief Scene Graph} with \textcolor[HTML]{00CC00}{\textit{blind nodes}} and to draw expectation about the location of an \textcolor[HTML]{FF0000}{artifact}, which enables a sequential high-level task planning
        .}
        \label{fig:diag}
        \vspace{-1.5em}
    \end{figure*}

To enable reasoning and planning in such a complex mission, 3D scene graphs (3DSG) 
have been introduced recently and are gaining popularity \cite{Chang2023}. In the context of 3D scene graphs, we can understand expectation as the likelihood of finding a specific object given a series of objects that have already been found. 
Substantial research has already been done in this field to improve and optimize the generation of 3D scene graphs, as well as for their implementation in a broad range of robotic tasks \cite{survey}. Nevertheless, even in the most recent state-of-the-art frameworks \cite{Kim2020,Wald2020,Wu2021,Wald2022,Feng2023,Wang2023,Wu2023,Rosinol2021,Hughes2022}, the concept of 3D scene graphs remains pretty much unchanged from when it was first proposed \cite{Armeni2019}. 
In contrast to previous works, in this article, we present an evolution of the classical concept, namely \textit{Belief Scene Graphs}, with the aim of enabling high-level reasoning for task planning and optimization of different robotic missions, given a partial representation of the environment in the form of a 3D scene graph. An overview of our approach is presented in Fig.~\ref{fig:diag}.



\subsection*{--- Related Work}

3D Scene graphs have emerged as a popular alternative for the encoding of scenes into intuitive and manageable data representations \cite{Kim2020,Wald2020,Wu2021,Wald2022,Feng2023,Wang2023,Wu2023}. 
3D scene graphs encode the environment in an abstract representation comprising of several layers (e.g. buildings, rooms and objects), in the form of nodes and edges, where nodes represent objects and edges represent the relationships among the nodes. 
Among the first works to propose the notion of 3D scene graphs is~\cite{Armeni2019}, where a semi-automatic framework for the construction of 3D scene graphs is presented. 
Since then, works such as ~\cite{Rosinol2021} and~\cite{Hughes2022} have proposed frameworks for the optimized construction of 3D dynamic scene graphs in more complex and large indoor environments. 

Nevertheless, it is only in the most recent state-of-the-art that we find the introduction of the idea of belief/expectation into 3D Scene Graphs, with the intention of enabling deeper reasoning with just partial information. Among these works we can find~\cite{Giuliari2022}, where a spatial commonsense graph is proposed in order to solve the problem of localization of objects in partial 3D scans. 
Furthermore, \cite{Looper2023} proposed a variable scene graph for semantic scene variability estimation, where three varieties of semantic scene change are identified: changes in position, semantic state, or scene composition.
Finally, \cite{ginting2023} presented a semantic belief graph for semantic-based planning under perceptual uncertainty to address navigation in extreme environments. 

\subsection*{--- Contributions}
In contrast to the state-of-the-art, the contributions of this article are summarized as follows. In this work, the focus is on extending the capability of a given partial 3D scene graph $\mathcal{G}^\prime$, in which objects of a set $C$ relevant for a robotic mission are missing, by proposing an approach to meaningfully incorporate $C$ as nodes into the scene graph $\mathcal{G}^\prime$. Towards this end, we propose a novel graph-based learning methodology, for the Computation of Expectation of finding objects in $C$ based on Correlation Information (CECI) between $C$ and the objects that are part of the given 3D scene graph $\mathcal{G}^\prime$. 
We design and implement a novel Graph Convolutional Network (GCN) model for CECI of 3D scene graphs. As no repository of 3D scene graphs is available for the training of the novel CECI model, we present a novel methodology for utilizing semantically annotated real-life 3D spaces to generate a 3D scene graphs dataset. The generated dataset is then utilized to train the proposed CECI model and also to validate the proposed methodology.
Secondly, we establish the novel concept of \textit{Belief Scene Graphs} (BSG), as a utility-enhanced extension of a given incomplete scene graph $\mathcal{G}^\prime$, by incorporating objects in $C$ into $\mathcal{G}^\prime$, using the learnt CECI information. \textit{Belief Scene Graphs} enable high-level reasoning and optimized task planning involving set $C$, which was impossible with the incomplete $\mathcal{G}^\prime$.
Finally, through extensive evaluation, we conclusively demonstrate the enhanced performance of the proposed BSG over classical 3DSG for the task of single-object and multi-object search and also present real-world results for prediction of unseen-objects by constructing a \textit{Belief Scene Graph} using information gathered by a legged robotic platform through the exploration of an unknown indoor environment. 
\section{PROBLEM FORMULATION}



\textbf{Modeling Expectation.} Expectation can be intuitively modeled as the conditional probability of an object to be found given other objects had already been found. 
%
%
Let $\mathcal{G} = (\mathcal{V}, \mathcal{E})$ denote a 3D Scene Graph, defined as a tuple of nodes $\mathcal{V}$ and edges $\mathcal{E}$. Similarly, let $C = \{c_1, ..., c_n\}$ with $n \in \mathbb{Z^+}$ denote the set of labeled object classes present in the graph $\mathcal{G}$. In this context, the probability for a given object class $c_i \in C$ to be found depends on a set of objects and room observations $B = \{b_1, ..., b_m\}$, $m \in \mathbb{Z^+}$, which are assumed to be independent of each other. Then the expectation for an object $c_i$ 
is given by 
$\boldsymbol{P(c_i|B)} = \sum_{j=1}^m P(c_i \cap b_j) / \prod_{j=1}^m P(b_j)$.
%
%
%
%
Note that the object $a_i$ and observation $b_j$ are being treated as single events with associated probabilities. 
Likewise, we can estimate the set of probabilities for finding every object class in the set $C$, that is:
\begin{equation} 
\boldsymbol{D =} 
\begin{pmatrix}
P(c_1|B) & P(c_2|B) & \cdots & P(c_n|B)
\end{pmatrix}.
\label{eqn:3} 
\end{equation}
Furthermore, many robotic applications also require information about the number of expected objects, which is encoded into the attribute set $A_B = \{a_{b1}, ..., a_{bm}\}$, with each element corresponding to an observation in set $B$. 
In this scenario, we need to estimate the conditional probability of the object $c_i$ with respect to the observations $B$ with $A_B$ attributes as:
\begin{equation} 
\boldsymbol{P(c_i|B, A_B)} = \sum_{j=1}^m P(c_i \cap b_j \cap a_{bj}) / \prod_{j=1}^m P(b_j \cap a_{bj})
\label{eqn:5} 
\end{equation}
Likewise, the conditional probability of an object class $c_i$ with $a_{ci} \in A_C$ attribute can be estimated through:
\begin{equation} 
\boldsymbol{P(a_{ci}|B, A_B)} = 
\sum_{j=1}^m P(a_{ci} \cap b_j \cap a_{bj}) / \prod_{j=1}^m P(b_j \cap a_{bj})
\label{eqn:6} 
\end{equation}
%

This reasoning is similar to the way humans draw expectations. For a human, the expectation to find an object $c_i$ in a given room is based on the nature of the object and the room itself
, the objects found in previous rooms (i.e. $B$) and their similarity with the expected object $c_i$. Even the nature of the building itself may temper the expectation of finding a specific object, for example, we will have lower expectations to find a pillow in an office building than in a residential building.

\textbf{Approximating Expectation}. Let $D_\mathcal{V}$ denote the real set of conditional probabilities of a set of labeled object classes $C$ with respect to all the nodes $\mathcal{V}$ of a given 3D scene graph. In real-life, translating the intuitive idea of expectations into a computational method for deriving $D_\mathcal{V}$ is challenging, due to the heavy data requirements for the purpose. To overcome this challenge, in this article, we look to find a reasonably approximate set of probabilities $E_\mathcal{V}$ for the set of probabilities $D_\mathcal{V}$, from the histograms of the set of nodes $\mathcal{V}$, where the histogram is equal to $E_\mathcal{V} \times O_\mathcal{V}$, where 
$O_\mathcal{V} \in \mathbb{Z}^+$ is the number of object nodes connected to each respective node in $\mathcal{V}$. Due to the sheer variability amongst real-life applications, the histogram of any given room is either unknown or hard to estimate for a field robot online. This is due to multiple factors, such as:
\begin{itemize}
\item \textit{Time}. The exploration is time-intensive or the robotic platform is time-constrained (e.g. drone flight time).
\item \textit{Blockades}. The physical limitations of the platform do not allow for the exploration of specific areas (e.g. a mobile robot reaching a steep change in elevation).
\item \textit{Occlusion}. The platform can only partially perceive the scene (e.g. a drone looking through a window).
\end{itemize}

\textbf{Computation of Expectation based on Correlation Information}. In order to be able to estimate $E_\mathcal{V}$ in partially known environments, and to allow the scalability of the framework, 
we propose a new model of Graph Convolutional Networks (GCN) for the Computation of Expectation base on Correlation Information (CECI) of 3D scene graphs. The novel CECI model aims to learn the underlying probability distribution for each node 
based on a dataset of sample environments. The CECI model should then be able to use sparse information about any given 3D scene graph topology, to draw expectations about possible objects present in the environment.
The next step is then to define the problem in terms of a GCN architecture, where the input is any given sparse 3D scene graph $\mathcal{G}^\prime$ with a set of histograms $E^{\prime}_\mathcal{N}(C) \times O^\prime_\mathcal{N}$ as node attributes over a set of labeled object classes $C$, where $\mathcal{N}$
is the set of nodes of interest, and the output is the set of histograms $E^{\prime\prime}_\mathcal{N}(C) \times O^{\prime\prime}_\mathcal{N}$ which can be used to approximate $\mathcal{G}^\prime$ to the ground truth graph $\mathcal{G}$ with a set of histograms $E_\mathcal{N}(C) \times O_\mathcal{N}$, as depicted in Fig.~\ref{fig:diag}.

%
\section{BELIEF SCENE GRAPHS}
\label{sec:ceci}
%
This section details the design process of the novel established CECI model, as well as the construction of the dataset for the training process. In addition, it presents the methodology for the generation of a \textit{Belief Scene Graph}.
\subsection{CECI Model Architecture Design and Optimization}
%
Depending on the diversity and scale of the environments, the number of nodes $\mathcal{V}$ may be substantial. In order to tackle this, 
the set of nodes $\mathcal{V}$ is masked to include only building and room nodes, which represent our set of nodes of interest $\mathcal{N}
\subset \mathcal{V}$. 
This is possible since the object nodes are already encompassed as node attributes in the set $\mathcal{N}$ (i.e. $E^{\prime}_\mathcal{N}(C) \times O^\prime_\mathcal{N}$). 
By doing so, we can optimize the learning process and the overall performance of the network.
Under these considerations, we propose a novel GCN architecture, namely CECI model. The proposed architecture consists of 5 GCN convolutional layers~\cite{gcn}, followed by batch normalization \cite{batchn}, ReLU and dropout. 
The input to the network is a given graph $\mathcal{G}^\prime$ with a set of histograms $E_\mathcal{N}^{\prime}(C) \times O^{\prime}_\mathcal{N}$ as node attribute and the output is the predicted set of histograms $E_\mathcal{N}^{\prime\prime}(C) \times O^{\prime\prime}_\mathcal{N}$, both with size $q \times n$, where $q \in \mathbb{Z}^+$ is the number of building and room nodes and $n \in \mathbb{Z}^+$ is the number of labeled object classes. The overall network architecture can be visualized in Fig.~\ref{fig:diag}. 
    
    
\subsection{Dataset Generation for CECI Model Training}
\label{sec:dataset}
Due to the novelty of the concept, currently there are no available datasets for the training of the CECI model. In order to generate the data needed for training, we used the Habitat-Matterport 3D Research Dataset (HM3D) \cite{hm3d}, currently the world's largest dataset of real-life residential, commercial, and civic spaces. First, we generated a custom-made mapping involving the grouping and filtering of the 1659 semantic categories present on the dataset, into a subset of 45 hand-picked object class labels selected based on their frequency on the scenes and their relevance to a robotic task. In addition, we have chosen to delimit the original set of room labels to a single general label (i.e. "room")~\cite{room,Rosinol2021,Hughes2022} to allow the proper generalization of the framework to unstructured indoor environments. This is an important consideration since the method is meant to work in partially known environments, where the semantic labels of the rooms are not always reliable. Then, for each of the dataset's semantically annotated 3D spaces, we constructed a 3D scene graph that includes 3 layers (i.e. building, rooms and objects) and the nodes consist of one attribute (i.e. the class label), while the edges represent descendant relationships (i.e. the child node is physically contained on the parent node). 

For each ground truth graph $\mathcal{G}$, we generated in addition a series of input graphs $\mathcal{G}^\prime$ by deleting object nodes at random, until the number of deleted nodes is equal to 50\% of the original number of nodes. The training data generation process was designed to resemble, as closely as possible, the information that a typical autonomous robotic system would acquire when deployed in a real-life mission and presented with different perception challenges (e.g. occluded objects, out-of-view objects, not visited areas, etc). 
Then the information of the remaining object nodes is encoded as a room node attribute, by encompassing the object nodes into a histogram of the 45 labeled object classes $C$ over the respective room node. Where the histogram represents the probability distribution $E_\rho^\prime(C) \times O^\prime_\rho$ over the room node $\rho \in \mathcal{N}$, for the defined set of objects $C$, where 
$O^\prime_\rho \in \mathbb{Z}^+$ is the number of object nodes connected to the room. Likewise, the ground truth histogram $E_\rho(C) \times O_\rho$ is estimated in the same way. In addition, the histograms $E_\eta(C) \times O_\eta$ and $E_\eta^\prime(C) \times O_\eta^\prime$ for the building node $\eta \in \mathcal{N}$ of every pair $\mathcal{G}$ and $\mathcal{G}^\prime$ is also computed. 

\subsection{Incorporating Expectation on Belief Scene Graphs}
\label{sec:bn}

The \textit{Belief Scene Graph} is defined as a tuple of nodes and edges $\mathcal{G}^{\prime\prime} = (\{\mathcal{B, \mathcal{V}}\},\mathcal{E})$, where $\mathcal{E}$ is the set of directional edges, $\mathcal{V}$ is the set of nodes presented in the original 3D scene graph $\mathcal{G}^{\prime}$, with the set of predicted histograms $E_\mathcal{N}^{\prime\prime}(C) \times O_\mathcal{N}^{\prime\prime}$, as node attribute for the building and room nodes, and $\mathcal{B}$ is the set of \textit{blind nodes}.
In a broad sense, a \textit{Belief Scene Graph} 
could be seen as an expansion of a 3D scene graph, through the incorporation of the concept of expectation. In this work, we represent expectation in two main ways, the first one is in the shape of the expected probabilistic distribution over a room node of the possible object nodes connected to it, and the second one is using a new kind of nodes called \textit{blind nodes}, which represent the expected objects to be found.

These nodes are generated in different ways, for example, we can complete the graph using the difference between the histograms of the input 3D scene graph  $E_\rho^\prime(C) \times O^\prime_\rho$ and the predicted one $E_\rho^{\prime\prime}(C) \times O^{\prime\prime}_\rho$ by adding connections from the room nodes to \textit{blind nodes} with the respective classes until the number of nodes and the class distribution of the room nodes matches the predicted one. On the other hand, we can also add only the \textit{blind nodes} of the labeled classes with the higher expectation corresponding to objects that were not originally present in the input 3D scene graph, 
but based on the current information of the environment are expected or \textit{believed} to be there.
The layer and information of these expectations depend on the specific task at hand, for example, in an exploration scenario, we might want to know the probabilities for certain zones to be found on unexplored paths, while in a pick-and-drop scenario, we may rather estimate the size and number of objects present in a set of pick-up locations. 

\section{EXPERIMENTAL EVALUATION}
In this section, we first provide the training parameters of the CECI model, followed by the validation using Wasserstein distance~\cite{wd}, energy distance~\cite{ed}, Mean Error (ME) and Frobenius norm~\cite{fn}, in addition, a qualitative analysis is also presented. Secondly, we present the simulation results for single-object and multi-object search. Finally, we report the results for the prediction of unseen-objects in a real-world indoor environment with a Boston Dynamics Spot Legged Robot.

\subsection{CECI Model Training}

In the developed implementation, we used a total of 45 labeled object classes. The generated dataset was used in an 80\% / 10\% / 10\% split for training, validation and testing respectively.
%
The training consisted of 5000 epochs with a batch size of 24. We used an Adam optimizer \cite{adam} with a learning rate of 0.01 and a learning rate decay of 5$e^{-6}$. 
The loss function was chosen to be Mean Squared Error (MSE).


\subsection{Validation Metrics}
During the validation, three main metrics were taken into consideration, the first one was the statistical distance between the probability distributions of the predicted histogram and the ground truth. For this we computed the Wasserstein distance (i.e. the earth mover’s distance) \cite{wd} and the energy distance~\cite{ed} 
for the pair of probabilistic distributions $E_\mathcal{N}^{\prime\prime}(C)$ for the prediction and $E_\mathcal{N}(C)$ for the ground truth. Table~\ref{table:distances} shows the mean, variance, skewness and kurtosis for each of the computed distances. Overall, it can be observed that the predicted distribution is fairly similar to the ground truth since it presents a low mean in both metrics in addition to a low variance.
\vspace{-0.5em}

\begin{table}[!htbp]
\caption{Statistical Distance}
\label{table:distances}
\centering
\begin{tabular}{c|c c c c}
\hline
\textbf{Metric} & \textbf{Mean} & \textbf{Variance} & \textbf{Skewness} & \textbf{Kurtosis} \\
\hline
\hline
Wasserstein & $0.0246$ & $8.2623e^{-5}$ & $-0.0840$ & $-0.6968$\\
Energy & $0.1122$ & $8.4677e^{-4}$ & $-0.8072$ & $0.4269$\\
\hline
\end{tabular}
\vspace{-1em}
\end{table}

\begin{figure}[!htbp]
        \centering
        \includegraphics[width=\columnwidth]{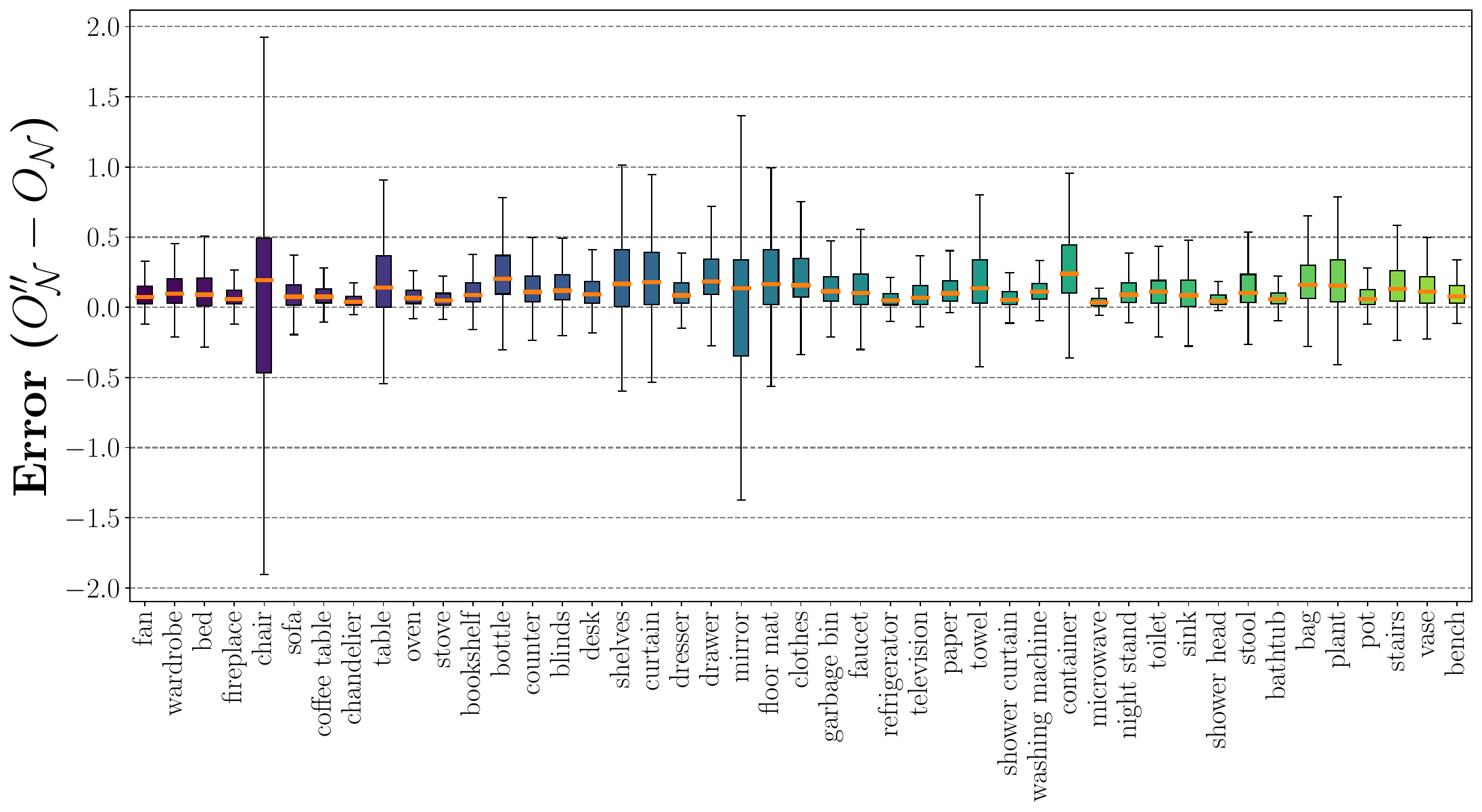}
        \caption{Comparison of the error between the predicted number of objects $O^{\prime\prime}_\mathcal{N}$ and the ground truth $O_\mathcal{N}$ per class label on the validation split.}
        \label{fig:box}
        \vspace{-0.5em}
    \end{figure}

The second metric was the error on the predicted number of objects $O^{\prime\prime}_\mathcal{N}$ over the set of labeled object classes $C$. Fig.~\ref{fig:box} shows the box plot for each of the classes, in particular, it shows how the common classes among multiple different kinds of rooms (e.g. chair, mirror, curtain, etc.) present high deviation and the higher mean error, while those specific to well-defined areas (e.g. chandelier, stove, refrigerator, bathtub, etc.) present low deviation and the lowest mean error.
    
\begin{figure*}[!ht]
        \centering
        \begin{subfigure}[b]{0.495\linewidth}
            \centering
            \includegraphics[width=\textwidth]{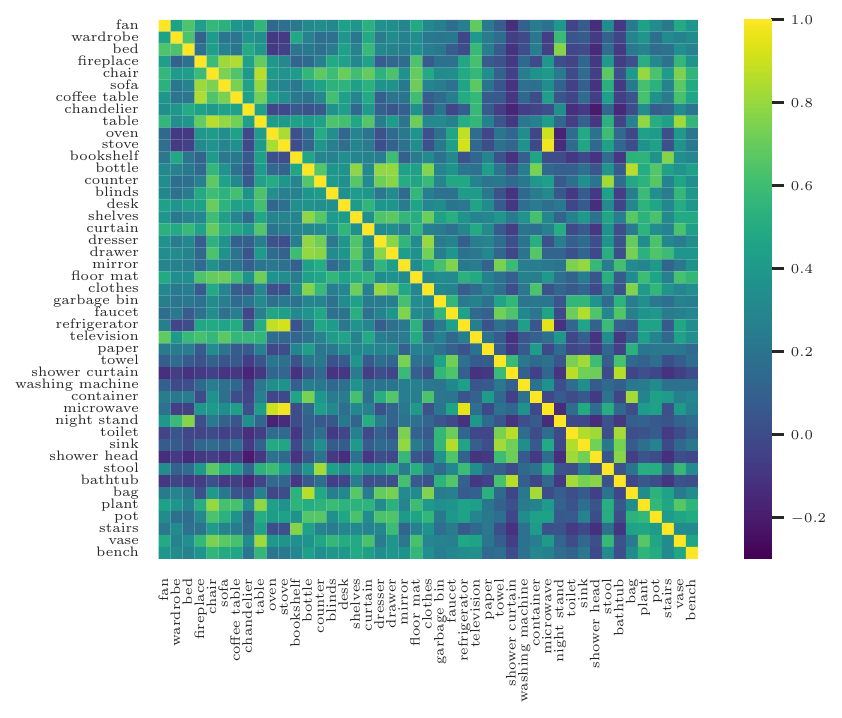}
            \caption{Predicted}
            \phantomsubcaption
        \label{fig:corr1}
        \end{subfigure}
        \begin{subfigure}[b]{0.495\linewidth}
            \centering
            \includegraphics[width=\linewidth]{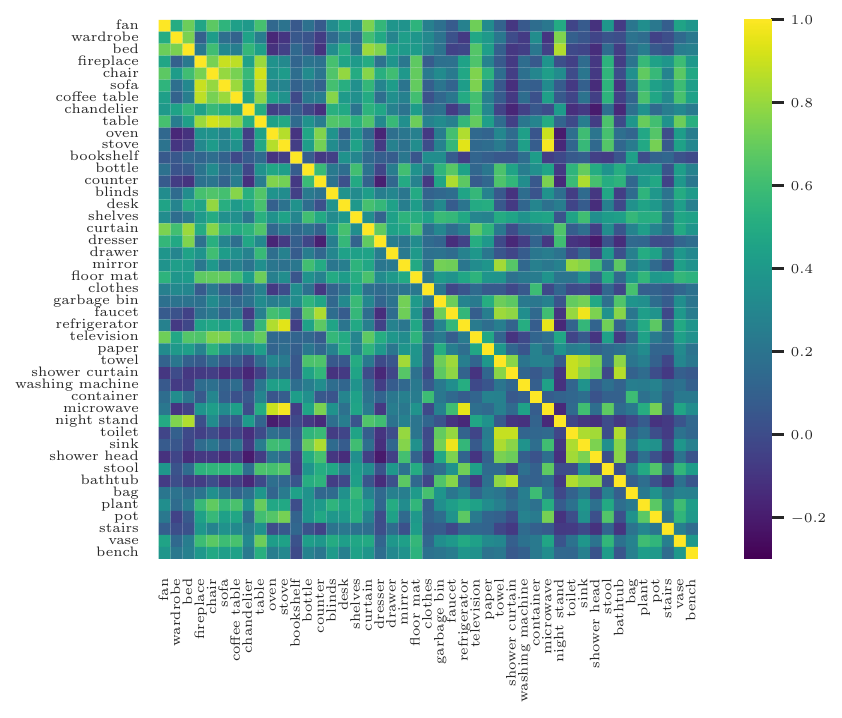}
            \caption{Ground Truth}
            \phantomsubcaption
        \label{fig:corr2}
        \end{subfigure}
        \caption{Comparison of the correlation matrices for the 45 labeled object classes of (a) the predicted graphs $\mathcal{G}^{\prime\prime}$ and (b) the ground truth $\mathcal{G}$ over the validation split.}
        \label{fig:corr}
        \vspace{-1.5em}
    \end{figure*}

Finally, we estimated the Frobenius norm \cite{fn} of the difference between the correlation matrices of the predicted and the ground truth data. The computed value 6.95 indicates a fair degree of similarity and showcases the performance of the CECI model for learning the underlying correlations between the different object class labels present in the generated data set. Fig.~\ref{fig:corr} presents both correlation matrices, where the main differences can be observed on the objects that are room-agnostic (i.e. can be found in multiple rooms, e.g. curtain and bottle). This behavior is consistent with Fig.~\ref{fig:box}.

\subsection{Qualitative Analysis}

\begin{figure*}[!ht]
        \centering
        \begin{subfigure}[b]{0.49\linewidth}
            \centering
            \includegraphics[width=\textwidth]{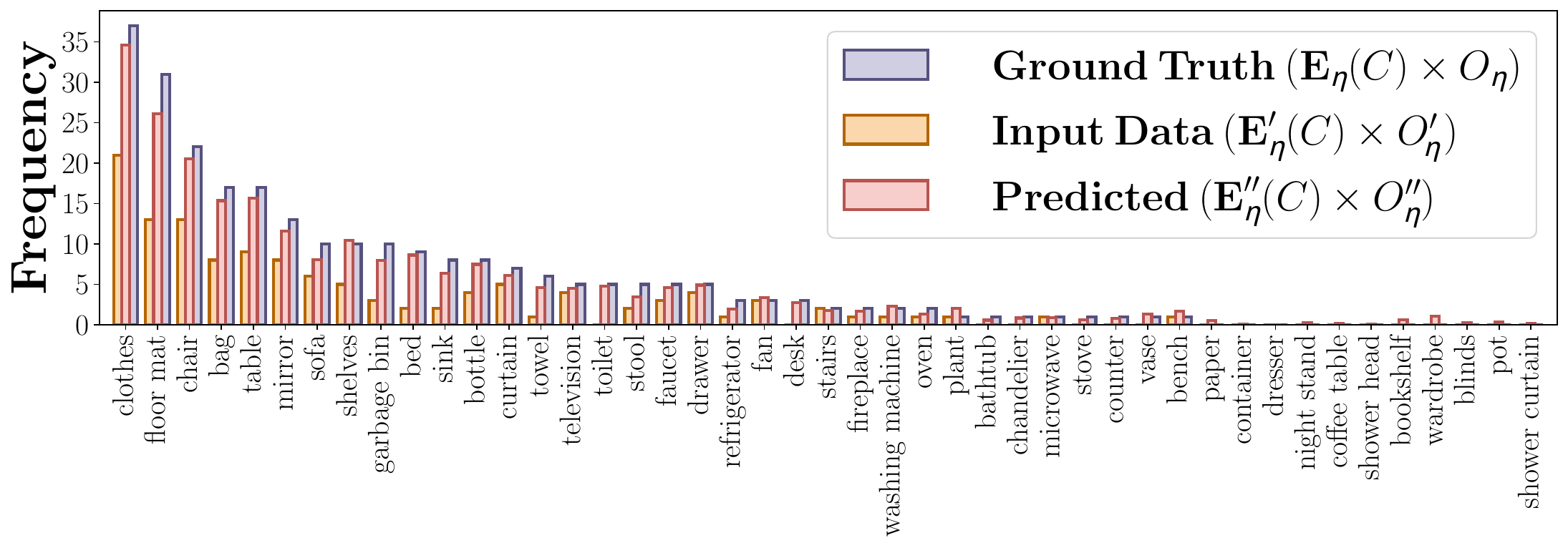}
            \caption{Histogram}
        \label{fig:hist}
        \end{subfigure}
        \hfill
        \begin{subfigure}[b]{0.49\linewidth}
            \centering
            \includegraphics[width=\linewidth]{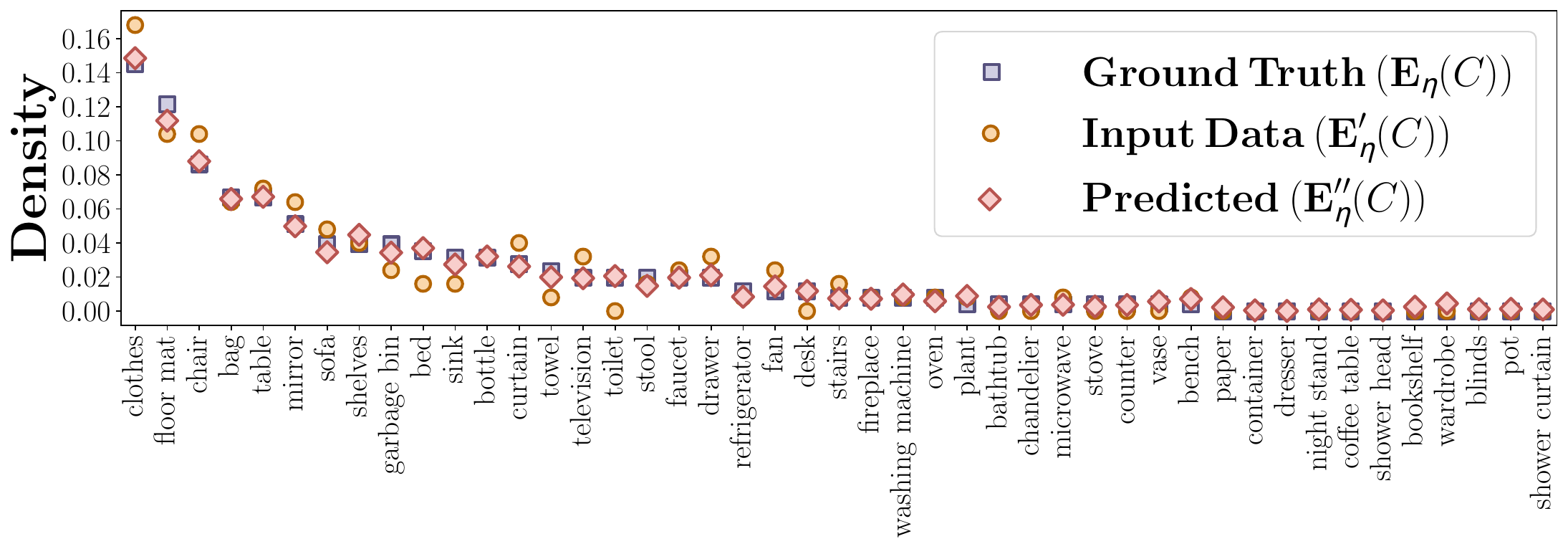}
            \caption{Probability Distribution}
        \label{fig:dist}
        \end{subfigure}
        \caption{Comparison of the input, predicted and ground truth (a) histograms and (b) the corresponding probability distributions over the building node of a single environment of the validation split.}
        \label{fig:qualitative}
        \vspace{-1.5em}
    \end{figure*}

The comparison between the input, predicted and ground truth histogram in Fig.~\ref{fig:hist} helps to illustrate the overall performance of the proposed model, in which the bigger differences are present at the object class labels with the higher frequency on the scene (e.g. clothes and floor mat), as well as in the ones with a frequency equal to 0 (e.g. bookshelf and wardrobe). More importantly, the capability of the model to predict objects that were not present in the input graph can be observed, for instance \textit{toilet} and \textit{desk}. This capability in particular is crucial and showcases the merit of the CECI method to compute expectation based on the correlation information of the graph taxonomy, which is one of the main contributions of this work.


The computed histogram can then be used to complete the 3D scene graph with \textit{blind nodes} as previously mentioned in~\ref{sec:bn}, as well as to extract the probability distribution $E_\mathcal{N}^{\prime\prime}(C)$ of the respective building or room node. 
Fig.~\ref{fig:dist} shows the distributions for the histograms presented in Fig.~\ref{fig:hist}, which helps to illustrate the results presented in Table~\ref{table:distances}. This is a key property to tackle the problem of object search efficiently, while different nodes may contain an instance of the artifact, the distribution of the room allows us to prioritize the ones that are more likely to contain it, in order to optimize the search base on the most likely node candidate. In other words, the one with the highest expectation of containing the artifact.

 


\subsection{Belief Scene Graphs for Object Search Task} \label{sec:simulation_results}


For the specifics of the experimental evaluation presented in this article, the application scenario will be centered around an object search task~\cite{Kurenkov2021,Ravichandran2022,Amiri2022} in simulation. The objective is to find an artifact or artifacts (i.e. a specific object class) in a partially known environment. This task was selected based on the high level of complexity for a variety of real-life robotic missions, like rescue operations, active source seeking or multi-agent task planning. 

\textbf{Object Search Definition:}
Given a partial representation of the environment, represented by a given graph $\mathcal{G}^\prime = (\mathcal{V},\mathcal{E})$, where $\mathcal{E}$ is the set of directional edges $e_{ij}(v_i,v_j) : (\mathcal{V} \times \mathcal{V}) \mapsto \mathcal{M}$ representing the cost of traversing from the node $v_i$ to the node $v_j$. In the task of object search, we need to find the optimal policy $\mathcal{P}^*$, which comprises of the set of edges to be traversed to reach the node $\mathcal{S} \in \mathcal{V}$ that contains the artifact $\alpha \notin \mathcal{V}$. 
For this task, the robotic system considers a graph $\mathcal{G}^\prime$ with traversal cost as edge attribute as follows:
\begin{itemize}
\item \textit{Building layer.} The evaluation scenario only considers single-building environments.
\item \textit{Rooms layer.} The layout of the rooms present in the environment, as well as the traversal cost $e$ associated with traversing through connected nodes. Nevertheless, the nodes don't contain specific class labels, 
in other words all nodes share the same class label (i.e. room). 
\item \textit{Objects layer.} A randomly selected subset of the objects present in the environment excluding any instances of the desired artifact $\alpha$. 
\end{itemize}

For comparison purposes, two algorithms were implemented, one using a 3D scene graph and the other using the proposed \textit{Belief Scene Graph}. More specifically,  Dijkstra’s graph search approach~\cite{dij} has been selected for the 3D Scene Graph, where the planner optimizes the traversal of the environment by prioritizing the shortest path to an unexplored room until the artifact is found. Regarding the \textit{Belief Scene Graph}, the graph search planner optimizes the traversal of the environment by prioritizing the room with the highest expectation of containing the artifact as $max(E_\rho^{\prime\prime}(\alpha)) \mapsto e_{ij}$.



Using the test split of the dataset, both algorithms were evaluated in a series of object search scenarios. The simulation runs were designed such that the position of the robot in the environment is randomized in every run, under the condition that it can not start in a room where the artifact is present. In addition, $e$ was randomly assigned for each simulated environment of the test split. The artifact is also switched to a different class label every certain number of runs. As such, the objective is for the robotic system to find an artifact or artifacts in a partially known environment with the lowest traversal cost. Once a room/area has been explored, the graph $\mathcal{G}^\prime$ will be updated with the rest of the objects present in that room. The task is finished once an instance of the artifact has been found for the first case scenario or once all the instances of the artifact are found for the second case scenario.

The simulation results showed that the total traversal cost computed over multiple randomized runs for the single-artifact case and 
the multi-artifact case is on average 33.8\% and 45.5\% lower, respectively, when using $\mathcal{G}^{\prime\prime}$ compared to $\mathcal{G}^{\prime}$. Demonstrating that the use of BSG-based planners yields better results overall for the task of object search compared to the classical 3DSG method.

\subsection{Field Test Results for CECI}

Additionally, to the simulation comparisons of the object search task presented in Section~\ref{sec:simulation_results}, we tested our approach on the Boston Dynamics Spot legged robot in a real civic indoor environment 
with the goal to contrast \textit{Belief Scene Graph} expectation with human common sense for unseen-objects. The environment consists of a small waiting room (i.e. Room A), a shared kitchen with a dining area (i.e. Room B), and a meeting room (i.e. Room C). The objective was for the robot to traverse and generate a 3D scene graph with partial information about the objects present in the scene, this graph was then enhanced using the proposed CECI model to draw expectations about other possible objects in the environment that were not part of the ones seen during the experiment. 

\begin{figure}[!htbp]
        \centering
        \includegraphics[width=1.0\linewidth]{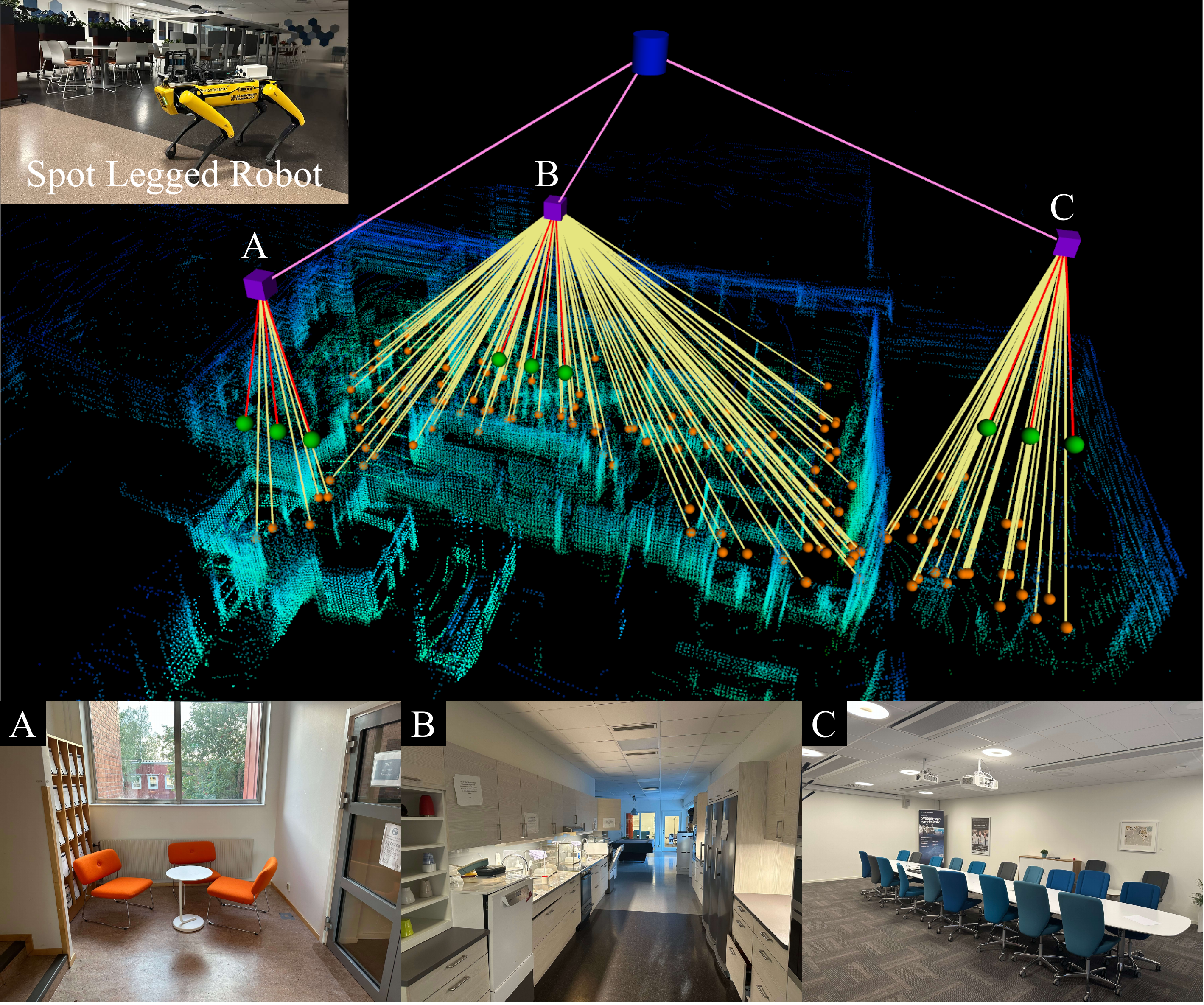}
        \caption{The \textit{Belief Scene Graph} generated using the spot legged robot in an indoor environment, where cylinders represent \textcolor[HTML]{0018EC}{buildings}, squares represent \textcolor[HTML]{9702fa}{rooms}, and spheres represent \textcolor[HTML]{FF8000}{objects} and \textcolor[HTML]{00CC00}{\textit{blind nodes}}.}
        \label{fig:fika}
        \vspace{-1.5em}
    \end{figure}

Figure \ref{fig:fika} presents the generated Belief Scene Graph of one of the runs, while depicting alongside images of each of the traversed rooms. 
In addition, Table~\ref{table:common} presents the top-3 most expected unseen-objects predicted on each of the rooms, i.e. those which present the higher expectation based on the histogram $E_\rho^{\prime\prime}(C) \times O^{\prime\prime}_\rho$, or $\underset{\text{top-3}}{max}(E_\rho^{\prime\prime}(U_\rho)) \mapsto \{u_{1^{st}}, u_{2^{nd}}, u_{3^{rd}}\}$, where $U_\rho \subset C$ is the subset of unseen-objects for the room $\rho \in \{A,B,C\}$. The results are presented with the following notation: $\diamond$ denotes those expectations that fall within the common sense and $\circ$ denotes expectations that fall within the common sense in the absence of context or under the presence of ambiguity.

\begin{table}[!htbp]
\setlength{\tabcolsep}{4pt} 
\centering
\caption{Top 3 Expectations for Unseen-objects}
\label{table:common}
\begin{tabular}{lr|lr|lr}
\hline
\multicolumn{2}{c|}{\textbf{Room A}} & \multicolumn{2}{c|}{\textbf{Room B}} & \multicolumn{2}{c}{\textbf{Room C }}  \\ 
\hline\hline
$\diamond$~plant       & 23.46\%     & $\diamond$ microwave    & 17.53\%    & $\circ$ vase     & 15.38\%            \\
$\diamond$~floor mat   & 10.74\%     & $\diamond$ sink         & 12.41\%    & $\circ$ fan      & 7.69\%             \\
$\diamond$ garbage bin & 5.27\%      & $\diamond$ refrigerator & 7.13\%     & $\diamond$ paper & 7.69\%             \\
\hline
\end{tabular}
\end{table}

\section{CONCLUSIONS}

A novel concept of \textit{Belief Scene Graphs} was introduced in this work, as a utility-driven extension of the classical 3D scene graphs. Reasoning about the presence/absence of new objects through expectations represents a crucial incremental step in the way robotics systems understand their environment, enabling close-to-human task planning and task optimization. The 
proposed CECI model (histograms learnt through GCN training) enabled the translation of an intuitive representation of beliefs into a computational method. The output histograms of the CECI model were then used to generate a \textit{Belief Scene Graph} by completing the 3D scene graph with \textit{blind nodes} and to extract the probabilistic distributions of the expected objects. The overall framework was experimentally validated for the task of object search. 
In addition, we present a field experiment in a real-world indoor environment showcasing the ability of the method to level with human common sense.
Future work will focus on the implementation of the framework on a broad range of robotic applications like exploration and task allocation for multi-agent scenarios.

\addtolength{\textheight}{-6.5cm}   








\bibliographystyle{ieeetr}
\bibliography{References}

\end{document}